\documentclass[10pt,twocolumn,letterpaper]{article}

\usepackage{cvpr} % To produce the CAMERA-READY version
\usepackage{fix-cm}
\usepackage[utf8]{inputenc} % allow utf-8 input
\usepackage[T1]{fontenc}    % use 8-bit T1 fonts
\usepackage{graphicx}
\usepackage{amsmath,bm}
\usepackage{amssymb}
\usepackage{booktabs}
\usepackage{enumitem}
\usepackage{acronym}
\usepackage{balance}
\usepackage{xspace}
\usepackage{setspace}
\usepackage{algorithm}
\usepackage{algpseudocode}
\usepackage{tabularx,colortbl,multirow,array,makecell}
\usepackage{overpic,wrapfig}
\usepackage{nicefrac}
\usepackage[misc]{ifsym}
\usepackage{siunitx}
\usepackage[dvipsnames,svgnames,x11names]{xcolor}
\usepackage{pifont}
\definecolor{cvprblue}{rgb}{0.21,0.49,0.74}
\usepackage[pagebackref,breaklinks,colorlinks,citecolor=cvprblue]{hyperref}
\usepackage[capitalize]{cleveref}
\usepackage{lipsum}

\newcommand{\supp}{\textit{supplementary}\xspace}

\definecolor{gblue}{HTML}{4285F4}
\definecolor{gred}{HTML}{DB4437}
\definecolor{ggreen}{HTML}{0F9D58}
\definecolor{vblue}{HTML}{2993ba}

\definecolor{gbest}{HTML}{FFCCCB}
\definecolor{gsecond}{HTML}{FFE5CC}
\definecolor{gthird}{HTML}{FFF2A0}
\newcommand{\modelname}{\textbf{\textsc{SCaR-3D}}\xspace}
\newcommand{\datasetname}{\textbf{\textsc{CCS3D}}\xspace}

\acrodef{git}[GIT]{Gaussian Instance Tracing}
\acrodef{nerf}[NeRF]{Neural Radiance Fields}
\acrodef{2dgs}[2DGS]{2D Gaussian Splatting}
\acrodef{3dgs}[3DGS]{3D Gaussian Splatting}
\acrodef{miou}[mIoU]{mean Intersection over Union}
\acrodef{macc}[mAcc]{mean Accuracy}
\makeatletter
\renewcommand{\paragraph}{%
  \@startsection{paragraph}{4}{\z@}%
  {1ex plus 0.5ex minus 0.2ex} % Space above (adjust this value)
  {-1em}                      % Space after the title
  {\normalfont\normalsize\bfseries} % Font styling
}
\makeatother

\def\paperID{332} % *** Enter the Paper ID here
\def\confName{3DV\xspace}
\def\confYear{2026\xspace}

\title{3D Scene Change Modeling With Consistent Multi-View Aggregation}
\author{
  Zirui Zhou\textsuperscript{1}\quad 
  Junfeng Ni\textsuperscript{1,2}\quad
  Shujie Zhang\textsuperscript{1}\quad
  Yixin Chen\textsuperscript{2,$\dagger$,\Letter}\quad
  Siyuan Huang\textsuperscript{2,\Letter}
  \\
  \textsuperscript{$\dagger$} Project lead \quad \textsuperscript{\Letter} Corresponding author \quad \\
  \textsuperscript{1} Tsinghua University \quad
  \textsuperscript{2} State Key Laboratory of General Artificial Intelligence, BIGAI\quad \\
  \url{https://zr-zhou0o0.github.io/SCaR3D/}
}

\begin{document}

\twocolumn[{
\renewcommand\twocolumn[1][]{#1}
\maketitle
\begin{center}
    \centering
    \captionsetup{type=figure}
    \includegraphics[width=\linewidth]{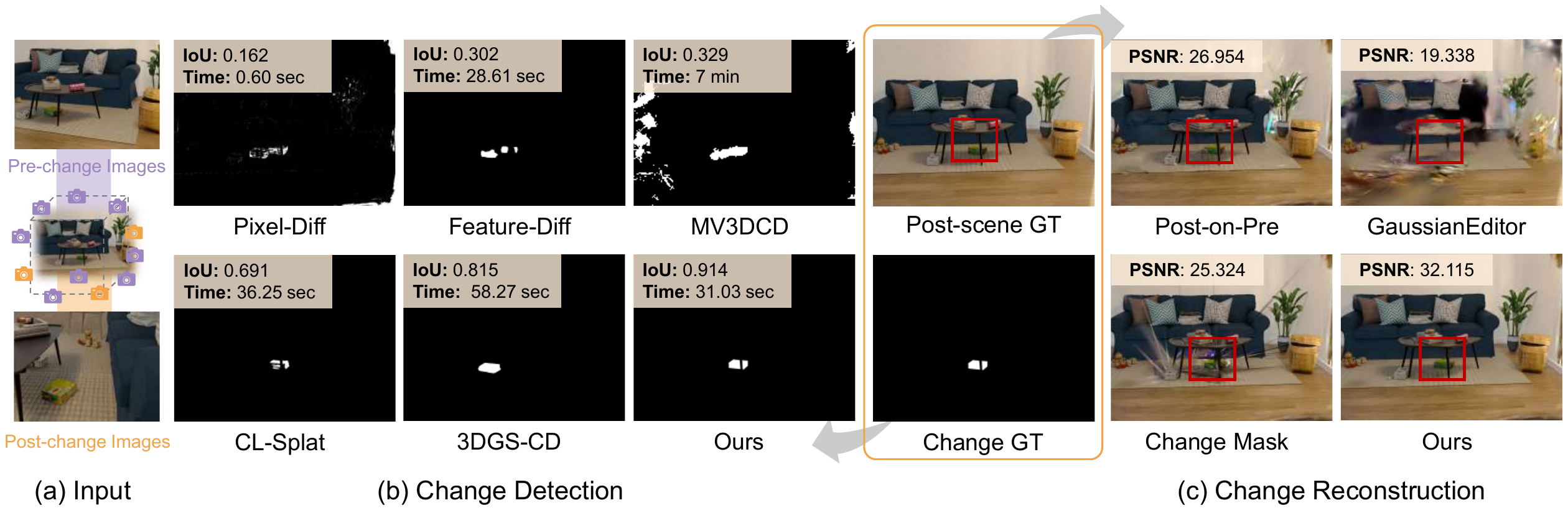}
    \captionof{figure}{
    We propose \textbf{\modelname}, a 3D scene change modeling framework that detects changes from dense-view pre-change images and sparse-view post-change images, while seamlessly reconstructing the post-change scene. \modelname significantly outperforms existing 3D change detection methods in change mask accuracy and computational efficiency, and delivers high-quality continual reconstructions. 
    }
    \label{fig:teaser}
\end{center}
}]

\begin{abstract}
Change detection plays a vital role in scene monitoring, exploration, and continual reconstruction. Existing 3D change detection methods often exhibit spatial inconsistency in the detected changes and fail to explicitly separate pre- and post-change states. To address these limitations, we propose \modelname, a novel 3D scene change detection framework that identifies object-level changes from a dense-view pre-change image sequence and sparse-view post-change images. Our approach consists of a signed-distance–based 2D differencing module followed by multi-view aggregation with voting and pruning, leveraging the consistent nature of 3DGS to robustly separate pre- and post-change states. We further develop a continual scene reconstruction strategy that selectively updates dynamic regions while preserving the unchanged areas.
We also contribute \datasetname, a challenging synthetic dataset that allows flexible combinations of 3D change types to support controlled evaluations. Extensive experiments demonstrate that our method achieves both high accuracy and efficiency, outperforming existing methods.

\end{abstract}

\section{Introduction}
\label{sec:intro}

% cliche: 3D reconstruction, applications
3D reconstruction~\cite{schoenberger2016sfm, yariv2021volume, wang2021neus, mildenhall2020nerf, kerbl3Dgaussians} is a fundamental task in computer vision, playing a crucial role in visual perception, embodied artificial intelligence (EAI), environment monitoring, and AR/VR~\cite{huang2024embodied, ni2024phyrecon, linghu2024msr3d, liu2025building, zhu2025mtu}. Real-world environments are inherently dynamic, where objects may appear, disappear, or translate and rotate over time. Much like a Sherlockian observer piecing together a scene from the smallest clues, a robust 3D reconstruction system must detect and interpret subtle environmental changes from the sparse, new observations through the lens of its 3D representation. Thus, reliable 3D change detection is essential to maintain an up-to-date and accurate representation of the evolving scenes. 

% what is change detection
% 2D change detection, challenges: noise; labeled; 
% previous 3D change detection
Change detection aims to identify objects in a scene that have changed between two time points, given multi-view images captured before and after the change. Previously, 2D change detection has been extensively studied, particularly in remote sensing applications such as monitoring land use changes, including the construction of buildings or roads~\cite{li2025dynamicearth}. However, these methods face significant limitations when applied directly to 3D scenes. First, most 2D approaches rely on supervised learning with annotated datasets, which are costly to create and often lack generalizability across different environments. Besides, these methods often struggle to maintain consistency across multiple views due to random noise and visual ambiguities, limiting their effectiveness in identifying coherent 3D changes.

% why 3dgs reconstruction
% why cd in 3dgs: view alignment, editable, consistency
Recent studies introduce 3D representations into change detection, with 3D Gaussian Splatting~(3DGS)~\cite{kerbl3Dgaussians} emerging as a particularly prominent approach. 3DGS enables efficient rendering of pre-change scenes from novel viewpoints in the post-change, and its explicit and editable structure facilitates the seamless identification and modification of the changed regions. Building on this, methods~\cite{ackermann2025clsplats,Galappaththige_2025_CVPR} encode change indicators directly into Gaussian primitives, yielding a unified representation of altered regions. However, such Gaussian-level representations lack holistic object-awareness, often producing fragmented change masks and view-dependent inconsistencies when representing the same object.
To mitigate these issues, 3DGS-CD~\cite{lu20253dgs} proposes identifying pre-change object masks using segmentation confidence, followed by learning pose transformations between pre- and post-change states. However, accurately matching masks of the same object consistently across multi-views remains challenging, leading to notable performance degradation under diverse change types such as translation, insertion, and removal.

% our method
Our method, Spotting Changes and Reconstruction in 3D Scenes (\modelname), is a multi-view voting-and-validation-based framework for efficient and consistent change detection in complicated and large-scale 3D scenes. Given two image sets captured from arbitrary viewpoints before and after scene changes, we first register their camera poses within a unified coordinate system. We then identify the feature-level difference between the pre-change and post-change observations by computing signed distance metrics. Utilizing a voting-based approach, we aggregate 2D differences from multiple perspectives,
and suppress noise and ensure geometric coherence via multi-view voting and pruning operations. The pruning strategy also robustly separates the pre-change and post-change difference.
Finally, we leverage EfficientSAM’s segmentation capability to validate the 3D differences and extract high-confidence change masks. By integrating these masks into the 3D reconstruction pipeline, our method enables continual reconstruction in regions where changes have occurred, while preserving the integrity of unchanged areas.

% experiment
% dataset
We introduce a new synthetic dataset tailored for 3D scene change detection, featuring complex and diverse indoor environments beyond simple tabletop settings. The dataset is fully editable, allowing flexible combinations of change types to support controlled evaluations. To assess the effectiveness of our method, we conduct experiments on both real-world datasets and synthetic datasets. Compared to existing methods, our approach produces more accurate and view-consistent change masks with higher efficiency.

In summary, our key contributions are as follows:
\begin{itemize} % [leftmargin=2em]  
    \item We propose a novel 3D scene change detection framework that leverages a 3D difference map and a multi-view consistency validation mechanism to accurately and efficiently identify object-level changes from two sequences captured under arbitrary viewpoints. 
    \item We construct a high-quality synthetic dataset, \datasetname, comprising editable indoor scenes for controlled evaluation of various 3D change types in complex environments.
    \item Extensive experiments demonstrate that our method outperforms previous approaches in terms of detection accuracy, change mask quality, and computational efficiency.
\end{itemize}

\section{Related Work}
\label{sec:related_work}

\subsection{Change Detection}
% 2D Change Detection
% 3D Change Detection
Change detection involves identifying regions or objects that exhibit differences by comparing images taken before and after the changes occur. 2D change detection from paired images has been a long-studied problem, with traditional methods such as~\cite{bovolo2006theoretical, celik2009unsupervised, luppino2019unsupervised}, and deep learning approaches~\cite{fang2022changer, dong2024changeclip, chen2021a, bandara2022transformer}.

Early 3D change detection methods, including image-based geometric approaches~\cite{taneja2011image, taneja2013city, sakurada2020weakly, palazzolo2017change, palazzolo2018fast} and TSDF-based detection~\cite{fehr2017tsdf}, provide important foundations but rely heavily on cadastral models or satellite-derived imagery, where pose and alignment errors are common. More recent geometric-consistency approaches also exhibit limitations in the types of changes they can handle: \cite{adam2022objects, lu20253dgs} assume that all changed objects appear in both pre- and post-change views to estimate rigid transformations, while \cite{langer2020robust} does not account for object removals.
  
 With the emergence of NeRF~\cite{mildenhall2020nerf} and 3DGS~\cite{kerbl3Dgaussians}, change detection can now operate on well-reconstructed scenes. For instance, \cite{huang2023c, martinson2024meaningful} train separate NeRFs on pre- and post-change images to detect changes from aligned views.\cite{lu20253dgs} aggregates 2D change masks into a 3D point cloud to learn pose changes, while \cite{Galappaththige_2025_CVPR} embeds change channels in 3DGS. Our method introduces an effective voting strategy to initialize a 3D difference map on 3DGS, validated by multi-view checks and segmentation confidence, enabling fast and accurate change detection.

% CL-Splat
\subsection{Continual Scene Reconstruction}
Continual 3D reconstruction aims to model a continuously updated 3D scene or its static background from an image sequence taken in dynamic environments~\cite{de2021continual}. However, directly training a scene representation over the sequence causes catastrophic forgetting~\cite{li2024learn, de2021continual} and degradation~\cite{wang2023sparsenerf}. To mitigate this, Li et al.~\cite{li2024learn} and Cai et al.~\cite{cai2023clnerf} introduce a keyframe database for historical image replaying. Another key topic for continual 3D reconstruction is to identify the transient regions to be excluded during model update. Traditional methods rely on depth residuals~\cite{palazzolo2019refusion} and pixel difference~\cite{fu2025gslts3dgaussiansplattingbased}. Learning based methods include \cite{li2024learn, cai2023clnerf, kulhanek2024wildgaussians} masking the transient objects with a learned classifier to maintain reconstruction consistency. Others~\cite{ackermann2025clsplats, kulhanek2024wildgaussians} exploits off-the-shelf vision model~\cite{caron2021emerging, xiong2023efficientsam} to identify the dynamic region. These methods highlight that effective change detection is essential for maintaining accurate and up-to-date 3D scene reconstructions over time, motivating our approach to integrate change detection with continual reconstruction.

\subsection{3D Editing}
3D editing refers to modifying specific parts of a reconstructed scene. Traditional 3D editing relies on human-operated tools such as Maya and Blender. For neural implicit representations such as NeRF and 3DGS, existing approaches primarily focus on text-driven~\cite{zhuang2024tip, GaussianEditor, palandra2024gsedit, wu2024gaussctrl, yan20243dsceneeditor, guo2024semantic, wang2021clip, ni2025dprecon} and image-based~\cite{zhuang2024tip, jaganathan2024ice, wang2021clip, bao2023sine} 3D editing. While existing methods provide stable edits and user-friendly interaction, they lack precise object insertion capabilities and depend on manual initiation. An alternative paradigm for 3D editing involves segmenting all objects in the scene, followed by selective editing of the targeted objects~\cite{gaussian_grouping, cen2025segment, guo2024semantic, gu2024egolifter, zhang2021stnerf, shen2025trace3d}. However, when only a few objects in a cluttered scene require editing, this approach leads to significant computational overhead. Our method leverages scene change detection to automatically trigger precise 3D edits, enabling efficient and targeted modifications by localizing updates to the detected change regions.

\section{Method}
% An overview of our method is illustrated in \cref{fig:pipeline}. Given two sets of images capturing the scene before and after changes, our goal is to detect object-level changes within a 3D scene. We first estimate their camera poses using COLMAP and render the pre-change scene from post-change viewpoints using 3DGS. We then apply EfficientSAM to compute signed distance for coarse 2D difference  (\cref{sec:2DDif}). These 2D differences are subsequently aggregated into a 3D representation through a multi-view voting and validation scheme. We further validate the 3D differences and extract high-confidence change masks by aligning them with the segmentation outputs from EfficientSAM (\cref{sec:3DDif}). Finally, the refined 2D change masks are used to perform localized updates to the 3D scene, enabling stable and accurate reconstruction in dynamic environments (\cref{sec:recon}).

\begin{figure*}[htbp]
    \hspace*{-0.2cm} % modify left or right
    \centering
    \includegraphics[width=5\linewidth, height=2.1in, keepaspectratio]{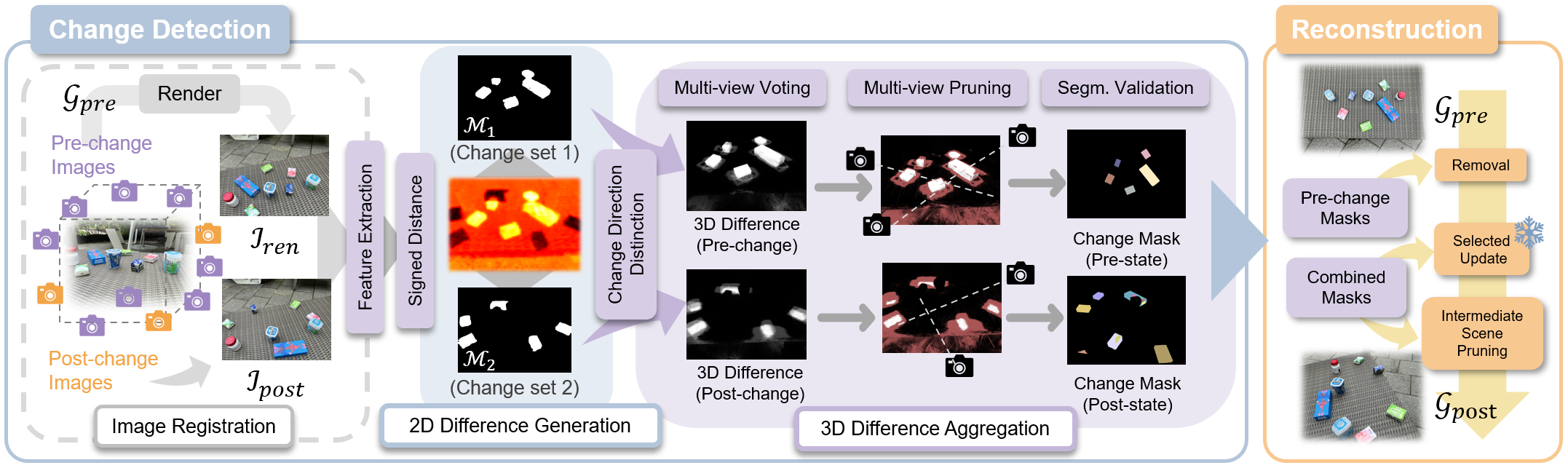}
    % \fbox{\rule{0pt}{2in} \rule{.9\linewidth}{0pt}}

    \caption{\textbf{Overview of \modelname.} 
    We first employ COLMAP for image registration, producing paired pre-change renders and post-change captures. 
    In the \emph{2D Difference Generation} stage, features are extracted and a signed distance metric is applied to separate the change regions into two sets. 
    After that, the \emph{3D Difference Aggregation} stage integrates multi-view differences through voting, pruning, and segmentation validation. 
    Finally, the change masks are applied to the reconstruction process to update the 3D scene selectively. }
    \label{fig:pipeline}
\end{figure*}

An overview of our method is shown in \cref{fig:pipeline}. Given pre-change and post-change image sets of a scene, we aim to detect object-level 3D changes. We first estimate camera poses and render the paired pre-post images (\cref{sec:ImgReg}). We then compute signed-distance maps for coarse 2D differences (\cref{sec:2DDif}), which are aggregated into 3D differences through multi-view voting and validation (\cref{sec:3DDif}). The resulting change masks then guide 3D updates, enabling stable and accurate continual reconstruction (\cref{sec:recon}). 

\graphicspath{{figures/}}

% formulation
% I: images
% M: masks O: objects 
% i_1, i_2...: n_pre pre-change view
% j_1, j_2...: n_post post-change view
% k_1, k_2...: n_test test view
% t_0 pre-change state
% t_1 post-change state
% O_1, O_2, ... O_n: n objects in the scene
% O_1^{t_0}: object 1 under state 0
% m_1, m_2, ... m_n: mask of n objects, under M
% {M_{t,o}^{i}} : mask of o_1, under state t_0, under view i_1

\subsection{Problem Setup}
\label{sec:problem_setup}
The input consists of two image sets:  $\mathcal{I}_{pre} = \{ I_i \mid i=1,\dots,n_{pre} \}$ captured from the pre-change scene under $n_{pre}$ arbitrary viewpoints, and $ \mathcal{I}_{post} = \{ I_i' \mid i=1,\dots,n_{post} \} $ from the post-change scene under $n_{post}$ viewpoints. We emphasize that $\mathcal{I}_{pre}$ represents a densely sampled set of views, whereas $\mathcal{I}_{post}$ corresponds to a sparsely sampled one. Our goal is to generate a set of change masks $ \mathcal{C} = \{ C_i \mid i=1, \dots, n_{test}\} $ under specified target viewpoints, and reconstruct the 3D scene $\mathcal{G}_{post}$ in 3DGS. 

\subsection{Image Registration}
\label{sec:ImgReg}

For $\mathcal{I}_{pre}$ and $\mathcal{I}_{post}$, we first leverage the structure-from-motion (SFM) algorithm~\cite{schoenberger2016sfm}, \eg, COLMAP~\cite{schoenberger2016mvs}, to simultaneously estimate their camera poses $\mathcal{P}_{pre}$ and $\mathcal{P}_{post}$. Assuming that the majority of scene features remain unchanged, we jointly register both image sets in a single SfM process to ensure that all estimated poses lie within a unified coordinate system. Additional implementation details are provided in \supp.  

We then train a 3DGS model using the pre-change image set $\mathcal{I}_{pre}$ to obtain a pre-change 3DGS $\mathcal{G}_{pre}$. We render $\mathcal{G}_{pre}$ from the post-change camera poses $\mathcal{P}_{post}$, producing $\mathcal{I}_{ren}$, where each rendered image in $\mathcal{I}_{ren}$ is paired with its corresponding real image in $\mathcal{I}_{post}$.  

\subsection{2D Difference Generation}
\label{sec:2DDif}  

\paragraph{Feature Extraction} We utilize EfficientSAM~\cite{xiong2023efficientsam} to extract image features $f$ from given image $I$:

\begin{equation}
{f} = \mathcal{F}{(I)},
  \label{eq:featExt}
\end{equation}
where $\mathcal{F}( \cdot )$ denotes the image encoder of EfficientSAM and $f \in R ^ {h \times w \times d}$. We bilinearly upsample EfficientSAM’s raw feature maps to the image resolution while keeping the embedding dimension $d$.

% We upsample the raw feature maps of EfficientSAM using bilinear interpolation to match the image resolution and preserve the original embedding dimension $d$. 

\paragraph{Signed Distance-Based Change Localization} To capture the directionality of changes, we adopt a signed distance formulation in the feature space. Specifically, for each feature map pair $(f_i, f_i')$, obtained from pairs of rendered pre-change image and post-change image via \cref{eq:featExt}, we apply Principal Component Analysis (PCA)~\cite{mackiewicz1993principal} to determine the dominant direction of variation. All pixel-wise feature vectors from $f_i$ and $f_i'$ are collected, and the first principal component vector $v$ is extracted as the direction of maximum variance. For each pixel $p$, the signed distance between pre- and post-change features is computed by projecting the feature difference onto $v$:

\begin{equation}
D_i^p = \frac{(f_i^p - f_i'^{p}) \cdot v}{\|v\|}.
  \label{eq:feat}
\end{equation}

Since the signed distance \(D_i^p\) separates foreground and background~\cite{lv2024weakly,dang2024beyond},
we threshold it to obtain two directional binary change masks:
\begin{align}
\mathcal{M}_{i,1} := \mathbf{1}\{\,D_i^p > \epsilon_1\,\}, 
\\
\mathcal{M}_{i,2} := \mathbf{1}\{\,D_i^p < \epsilon_2\,\}.
\label{eq:sd}
\end{align}
\(\mathbf{1}\{\cdot\}\) is indicator function, and \(\epsilon_1 \ge 0 \ge \epsilon_2\) are thresholds.

\graphicspath{{figures/}}
\begin{figure}
    \centering
    \includegraphics[width=\linewidth, height=2.8in, keepaspectratio]{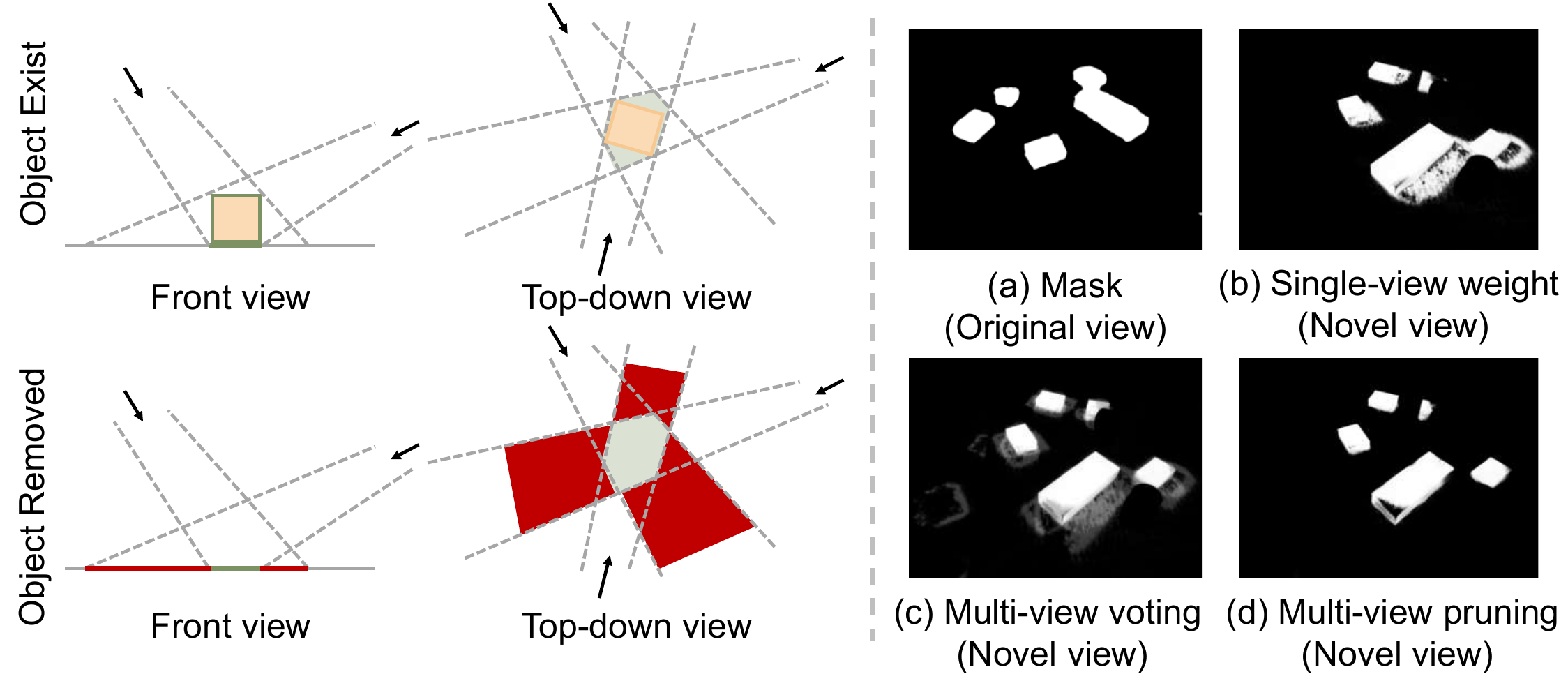}
    \caption{\textbf{Multi-view pruning.} Left: When objects exist (top), masks are scattered and few Gaussians are removed; when objects are removed (bottom), background tracing leads to extensive pruning. Right: Novel-view 3D difference visualization for (b) single-view weighting in ~\cref{eq:trace_sing}, (c) multi-view voting in ~\cref{eq:trace_seen}, and (d) multi-view pruning.}
    \label{fig:3DDif}
\end{figure}

\subsection{3D Difference Aggregation}
\label{sec:3DDif}

Although the 2D difference masks $\mathcal{M}_{i,1}$ and $\mathcal{M}_{i,2}$ incorporate image-level semantic features and are more robust than raw pixel-level comparisons, they still suffer from noise and viewpoint-induced ambiguities. To mitigate these issues, we aggregate the 2D differences into a unified 3D representation, leveraging spatial consistency across multiple views.

\paragraph{Multi-view Voting} To aggregate the 2D differences into 3D, we initialize a 3D difference representation based on the pre-trained pre-change 3DGS model $\mathcal{G}_{pre}$. Specifically, we embed an additional difference channel into each Gaussian to indicate whether it has changed.

For every single view, following the semantic tracing method introduced in GaussianEditor~\cite{GaussianEditor}, we identify and update the relevant Gaussians for each 2D mask by computing their contribution: 

\begin{equation}
w_i = \sum_{p} o_i(p) \cdot T_i(p) \cdot M(p) ,
\label{eq:trace_sing}
\end{equation}
where $w_{i}$ represents the weight of the $i$-th Gaussian, $o_i(p)$, and $T_i(p)$ denote the Gaussian's opacity, transmittance from pixel $p$, and $M(p)$ the 2D mask of pixel $p$. To normalize the weights, we define $\tilde{w}_i = \frac{w_i}{w_{\max}}$, where $w_{\max}$ is the maximum weight across all Gaussians in the current view, ensuring that $\tilde{w}_i \in [0,1]$.

For the multi-view setting, a straightforward approach is to aggregate weights from all post-change views. Let $S_i^k = \sum_{p} o_i^k(p) \, T_i^k(p) \, M^k(p)$ denote the aggregated contribution of the $i$-th Gaussian from the $k$-th view. A simple normalization by the total number of post-change views is:

\begin{equation}
w_i = \frac{1}{n_{\text{post}}} \sum_{k=1}^{n_{\text{post}}} \frac{S_i^k}{w_{\max}^k}
\label{eq:trace_vote}
\end{equation}

However, this uniform normalization by $n_{\text{post}}$ introduces bias against Gaussians visible in fewer views due to occlusions or restricted fields of view, assigning them disproportionately low weights. To address this, we adopt a visibility-aware strategy, normalizing each Gaussian's weight by the actual number of views in which it is observed, $n_i^{\text{seen}}$:
\begin{equation}
w_i = \frac{1}{n_i^{\text{seen}}} \sum_{k=1}^{n_i^{\text{seen}}} \frac{S_i^k}{w_{\max}^k}.
\label{eq:trace_seen}
\end{equation}

% To visualize the aggregated 3D difference map, the original view-dependent colors represented by spherical harmonics are replaced during rendering with these computed difference weights, which are mapped from a normalized [0, 1] range to [0, 255] grayscale values, as shown in \cref{fig:3DDif}.

Our method leverages the fast GPU Radix sort algorithm already implemented in the 3DGS rendering pipeline, enabling us to process a scene in just a few seconds, whereas methods such as GaussianCut~\cite{jain2024gaussiancut} require several minutes per scene and incur additional memory overhead.

\graphicspath{{figures/}}
\begin{figure*}[htbp]
    \centering
    \includegraphics[width=\linewidth, height=30in, keepaspectratio]{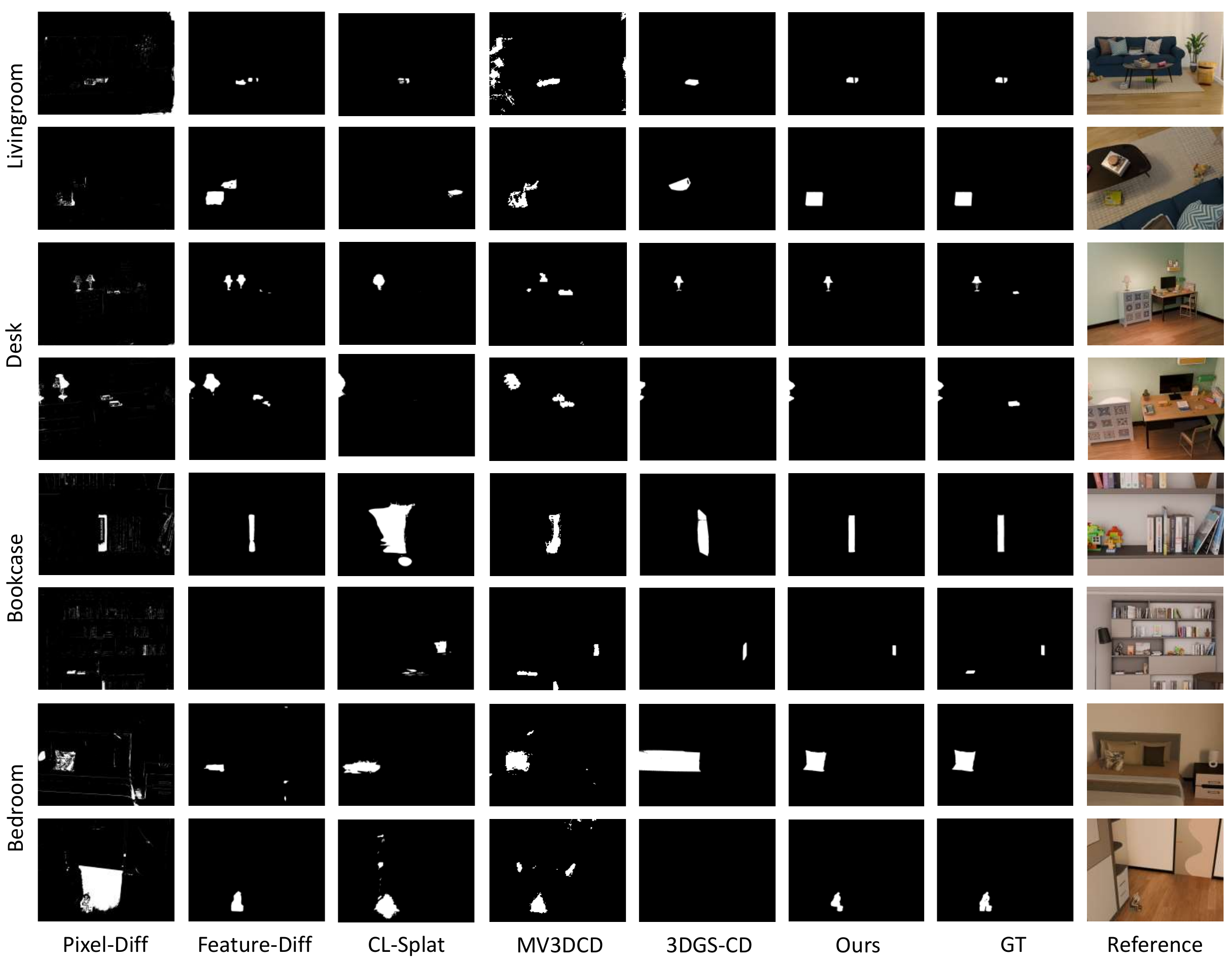}
    % \fbox{\rule{0pt}{2in} \rule{.9\linewidth}{0pt}}
    \caption{\textbf{Qualitative change detection results on the \datasetname dataset.} Each pair of rows corresponds to a single scene captured from different viewpoints. The last column, labeled \emph{Reference}, shows the post-change images from the matched viewpoints.}
    \label{fig:result_blender}
\end{figure*}

\paragraph{Multi-view Pruning} We observe from \cref{fig:3DDif} that the accumulated weights lack object-level awareness. In some cases, the weights erroneously bleed through foreground objects and are projected onto the background, leading to inconsistent aggregation.
To mitigate this issue, we perform a multi-view consistent pruning step after the voting process. Specifically, we remove Gaussians whose centers are projected outside the 2D masks in more than $\tau$ out of the total $n_{\text{post}}$ views. 

% Let $\mu_i$ be the center of the $i$-th Gaussian in 3D space. $\pi_k(\mu_i)$ denotes its projection onto the $k$-th view, yielding a pixel location $p_k^i$. Then, define an indicator function:

% \begin{equation}
% \delta_k^i =
% \begin{cases}
% 1, & \text{if } M^k(p_k^i) = 0 \\
% 0, & \text{if } M^k(p_k^i) = 1
% \end{cases}
% \label{eq:pruneIndc}
% \end{equation}

% Here, $M^k(p)$ is the 2D mask value at pixel $p$ in view $k$. We then count how many views consider the Gaussian center as outside the mask:

Let $\mu_i$ denote the center of the $i$-th Gaussian in 3D space, and $p_k^i$ be its projection, \ie, the pixel location, onto the $k$-th view. For each view $k$, we define an indicator $\delta_k^i = 1$ if the 2D mask value $M^k(p_k^i)$ at pixel $p_k^i$ is zero (\ie, outside the mask), and $\delta_k^i = 0$ otherwise. The total number of views in which the Gaussian center lies outside the mask is:
\begin{equation}
n_{\text{out}}^i = \sum_{k=1}^{n_i^{seen}} \delta_k^i.
\label{eq:pruneSum}
\end{equation}
Finally, the $i$-th Gaussian is pruned if $n_{\text{out}}^i > \tau \cdot n_i^{seen}$.

\paragraph{Change Direction Distinction} As introduced in \cref{sec:2DDif}, we use signed distances to separate change regions into two directional categories, but we cannot yet determine which category indicates object addition and which indicates object removal. Since added objects can only be found in post-change scenes, and removed objects can only be found in pre-change scenes, we combine the multi-view pruning strategy in \cref{sec:3DDif} with the following procedure to further infer whether each directional change mask corresponds to objects in the pre-change scene or the post-change scene.

Let $N$ denote the total number of Gaussians selected during the voting process, and let $N_p$ denote the number of Gaussians pruned during the consistency check. We define the retention rate as $R = \frac{N - N_p}{N}$. As illustrated in \cref{fig:3DDif}, when an object exists at the hypothesized location, the projected mask aligns well with the actual object surface, resulting in a higher weight concentration and a higher retention rate $R$. Conversely, when the object is absent, weights are more likely to scatter and get pruned, leading to a lower $R$. Applying 3D difference tracing to the pre-change Gaussians ($\mathcal{G}_{pre}$) associates a higher retention rate ($R$) with the pre-change state and a lower rate with the post-change state.

% Since we apply 3D difference tracing on pre-change Gaussians $\mathcal{G}_{pre}$, a directional change mask with a higher retention rate $R$ is thus associated with the pre-change state, while the direction with a lower $R$ corresponds to the post-change state.

\begin{table*}[ht]
\centering
\caption{\textbf{Quantitative change detection results on the \datasetname dataset.} The best, second-best, and third-best scores are highlighted in red, orange, and yellow, respectively. Our method demonstrates the best overall performance.}
\begin{tabular}{l c c c c c c c c c c}
\toprule
& \multicolumn{2}{c}{Livingroom} & \multicolumn{2}{c}{Desk} & \multicolumn{2}{c}{Bookcase} & \multicolumn{2}{c}{Bedroom} & \multicolumn{2}{c}{Average} \\
\cmidrule(lr){2-3} \cmidrule(lr){4-5} \cmidrule(lr){6-7} \cmidrule(lr){8-9} \cmidrule(lr){10-11}
Method & F1 & IoU & F1 & IoU & F1 & IoU & F1 & IoU & F1 & IoU \\
\midrule
% \cmidrule(lr){1-1} \cmidrule(lr){2-3} \cmidrule(lr){4-5} \cmidrule(lr){6-7} \cmidrule(lr){8-9} \cmidrule(lr){10-11}
Pixel-Diff    & 0.273 & 0.162 & 0.398 & 0.254 & 0.315 & 0.201 & 0.286 & 0.176 & 0.318 & 0.198 \\
Feature-Diff   & 0.420 & 0.302 & 0.480 &  0.323 & 0.320 & 0.256 & \cellcolor{gsecond} 0.705 & \cellcolor{gsecond} 0.584 & 0.450 & 0.343 \\
CL-Splat  &  \cellcolor{gthird} 0.789 & \cellcolor{gthird}0.657 & \cellcolor{gsecond}0.567 & \cellcolor{gthird} 0.399 & 0.294 & 0.199 & 0.501 & 0.341 & \cellcolor{gsecond} 0.538 & \cellcolor{gthird}0.399   \\
MV3DCD    & 0.478 & 0.329 & 0.291 & 0.178 & \cellcolor{gsecond} 0.449 & \cellcolor{gthird} 0.295 & \cellcolor{gthird} 0.547 & \cellcolor{gthird} 0.413 & 0.441 & 0.304 \\
3DGS-CD   & \cellcolor{gsecond} 0.897 &  \cellcolor{gsecond}0.815 & \cellcolor{gthird} 0.525 & \cellcolor{gsecond} 0.408 & \cellcolor{gbest} 0.477 & \cellcolor{gsecond} 0.353 & 0.148 & 0.089 & \cellcolor{gthird} 0.512 & \cellcolor{gsecond} 0.416 \\
Ours      & \cellcolor{gbest} 0.955 & \cellcolor{gbest} 0.914 & \cellcolor{gbest} 0.610 & \cellcolor{gbest} 0.477 & \cellcolor{gthird}0.423 & \cellcolor{gbest} 0.377 & \cellcolor{gbest} 0.909 & \cellcolor{gbest} 0.834 & \cellcolor{gbest} 0.724 & \cellcolor{gbest} 0.650 \\
\bottomrule
\end{tabular}
\label{tab:blender}
\end{table*}

\paragraph{Segmentation Validation} To estimate the reliability of the 3D difference map and refine it into a high-quality 2D mask, we validate the difference against predictions from EfficientSAM. For each view, we simply select the EfficientSAM mask with the highest IoU against the projected 3D difference.

% For each view, we compute the Intersection-over-Union (IoU) between the projected 3D difference and all masks generated by EfficientSAM in its segment-everything mode. If the maximum IoU exceeds a predefined threshold $\tau$, the corresponding EfficientSAM mask is considered valid and selected as the final segmentation result.

\subsection{Continual Scene Reconstruction}
\label{sec:recon}

Updating $\mathcal{G}_{pre}$ to $\mathcal{G}_{post}$ presents several challenges. First, given the sparsity of post-change views, it is crucial to prevent degradation of Gaussians that are unseen in the post-change images. A straightforward approach of mixing post-change images with pre-change images mitigates this issue, but it significantly increases computational cost and introduces view-dependent ambiguities within the changed regions. Alternatively, freezing Gaussians outside the change masks is a common strategy; however, insufficient supervision near the boundaries of the 2D change masks often leads to drifting Gaussians along the edges of changed areas.

To mitigate these problems, we adopt a 2D change mask–based strategy that replaces pre-change objects with their post-change counterparts during reconstruction. Specifically, we first apply the pre-change masks to $\mathcal{G}_{pre}$ to remove the corresponding objects. Then, we use the masked post-change images $\mathcal{I}_{post}$ to locally update $\mathcal{G}_{pre}$, producing an intermediate scene. In this process, Gaussians outside the 2D change masks are frozen, while loss computation and gradient descent are restricted to the masked regions in screen space. Finally, leveraging the voting and pruning method introduced in \cref{sec:3DDif}, we replace the change region in  $\mathcal{G}_{pre}$ with the corresponding regions from the intermediate scene. This results in the updated post-change scene $\mathcal{G}_{post}$ with fewer drifting artifacts from view-dependent degradation while accurately relocating the changed objects to their new position. 
% The localized update approach further improves reconstruction fidelity within the changed regions, as it may circumvent sub-optimal results.

% As shown in \cref{fig:Recon}, our method yields fewer drifting artifacts from view-dependent degradation while accurately relocating the green box to its new position. The reconstruction results on post-change views (\cref{fig:ReconPost}) further indicate that our local optimization strategy achieves higher fidelity in the changed regions.

%---------------------------------------------------------------------/

\section{Experiments}
\label{sec:experiment}

\subsection{Experiment Setup}
\label{sec:expSetup}

% \begin{table}[ht]
% \centering
% \caption{\textbf{Evaluation of our method on the Livingroom scene (\datasetname).} The first column lists the change types present in the scene. The second column shows the number of changed objects.}
% \label{tab:detailed_results}
% \begin{tabular}{lccccc}
% \toprule
% \textbf{Change Type} & \textbf{\# Obj} & \textbf{Precision} & \textbf{Recall} & \textbf{F1} & \textbf{IoU} \\
% \midrule
% In and Out  & 1 & 0.961 & 0.947 & 0.953 & 0.911 \\
% In and Out  & 2 & 0.782 & 0.758 & 0.770 & 0.742 \\
% In and Out  & 4 & 0.782 & 0.748 & 0.764 & 0.732 \\
% Translation & 1 & 0.992 & 0.978 & 0.985 & 0.970 \\
% Translation & 2 & 0.794 & 0.501 & 0.613 & 0.498 \\
% Translation & 4 & 0.805 & 0.540 & 0.585 & 0.472 \\
% Rotation    & 1 & 0.999 & 0.965 & 0.982 & 0.964 \\
% Rotation    & 2 & 0.798 & 0.613 & 0.690 & 0.613 \\
% Rotation    & 4 & 0.798 & 0.253 & 0.383 & 0.253 \\
% Mixed       & 2 & 0.597 & 0.458 & 0.390 & 0.307 \\
% Mixed       & 4 & 0.994 & 0.260 & 0.408 & 0.260 \\
% \bottomrule
% \end{tabular}
% \label{tab:Control_dataset}
% \end{table}

\graphicspath{{figures/}}
\begin{figure*}
    \centering
    \includegraphics[width=\linewidth, height=5.7in, keepaspectratio]{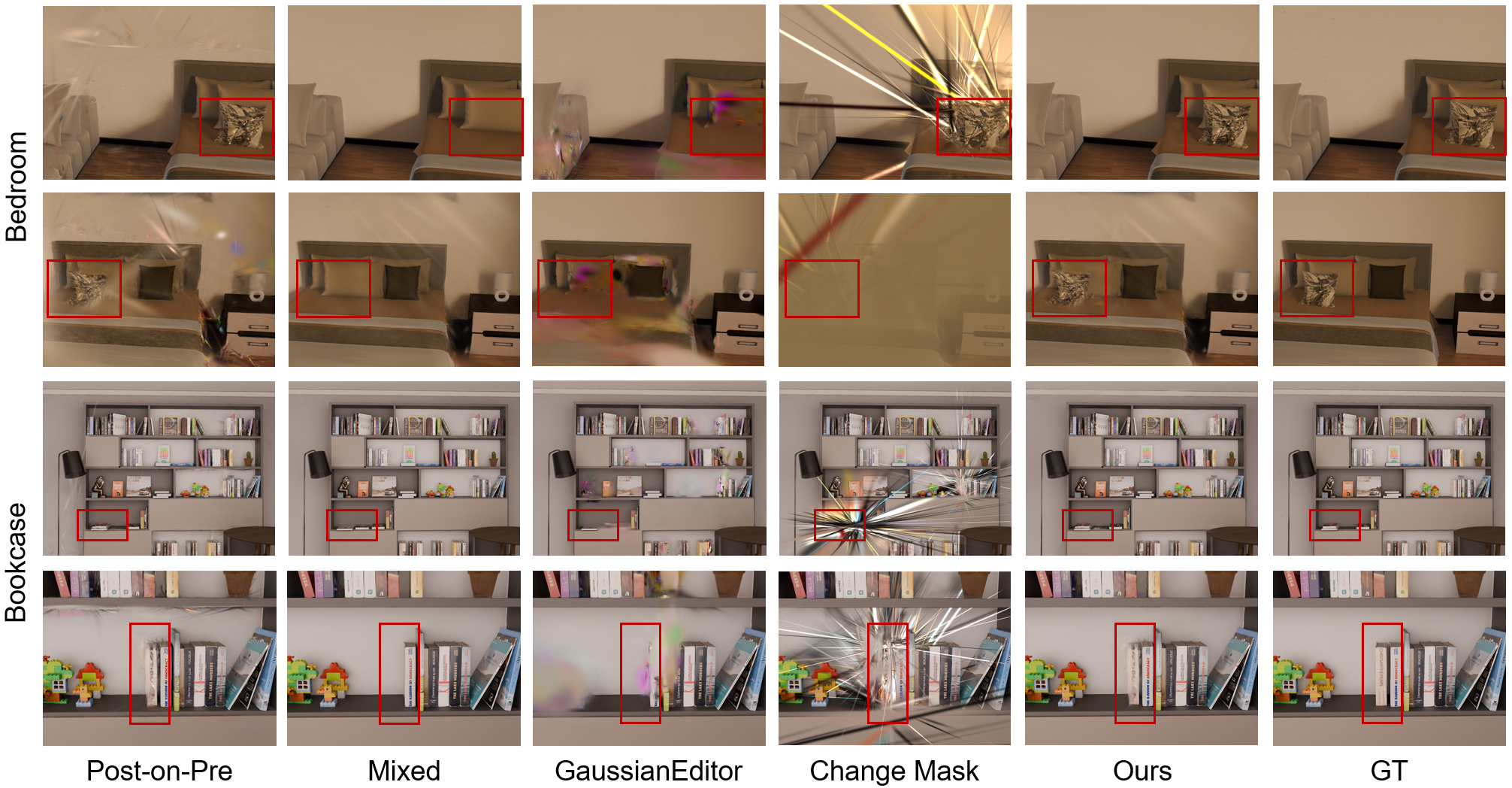}
   \caption{\textbf{Qualitative reconstruction results on novel views.} Each pair of rows corresponds to a single scene captured from different novel
viewpoints. The changed regions are highlighted with red boxes.}
    \label{fig:Recon}
\end{figure*}

\graphicspath{{figures/}}
\begin{figure}
    \centering
    \includegraphics[width=\linewidth, height=2.8in, keepaspectratio]{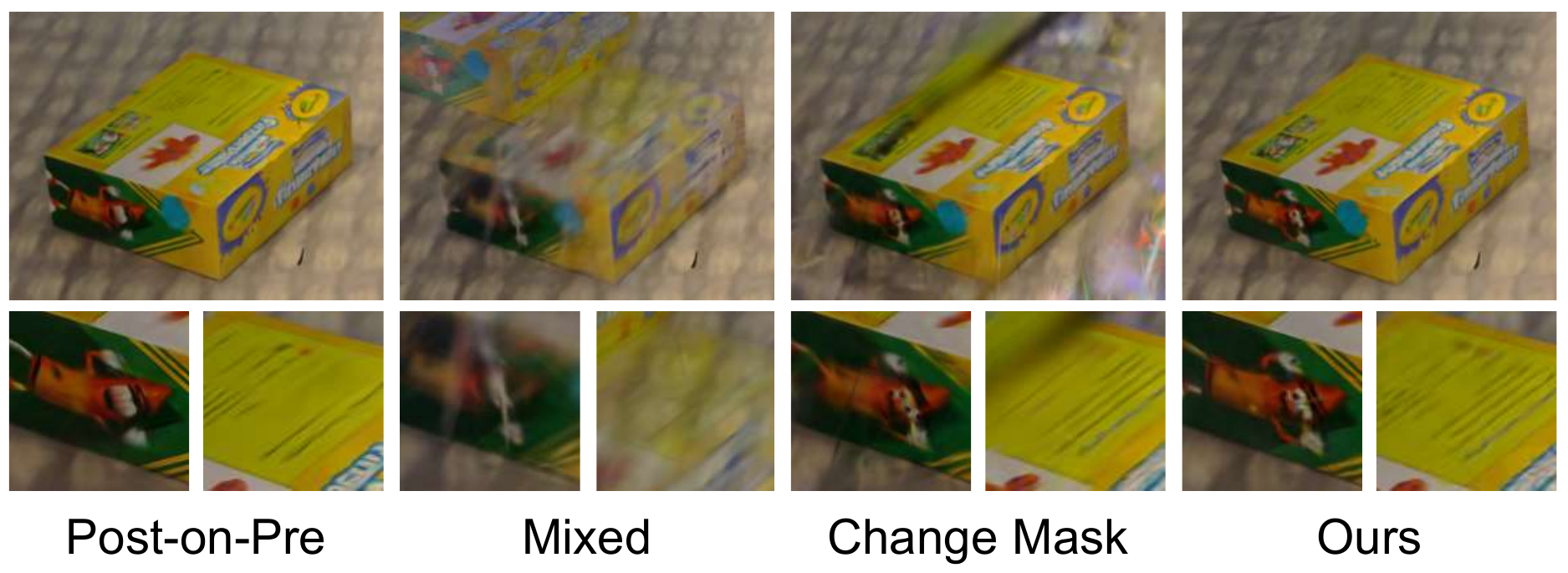}
    \caption{\textbf{Reconstruction results on post-change views.}}
    \label{fig:ReconPost}
\end{figure}

\paragraph{Datasets}
We first introduce a new dataset, Controllable Change in 3D Scenes (\datasetname), which comprises four diverse and comprehensive synthetic scenes: Desk, Bookcase, Livingroom, and Bedroom, constructed with Blender~\cite{blender}. Compared with existing 3D change detection datasets, \datasetname is not restricted to tabletop scenarios with simple camera trajectories (\eg, face-forward or fixed 360° rotation). The Bookcase scene features a multi-floor bookshelf, where the camera navigates from a distant view to a close-up inspection, sequentially exploring each shelf. The Livingroom and Bedroom scenes offer complete 360° environments containing both large-scale furniture (\eg, chairs, tables) and small tabletop items (\eg, books, pencil cases), designed to support fine-grained change detection. They incorporate complex, human-like camera navigation patterns, simulating natural walking and exploration. Furthermore, our dataset enables controlled experiments on change detection with varying numbers of objects and change types, where such control is difficult to achieve in existing 3D change detection datasets.

We also evaluate our model on the real-world dataset, 3DGS-CD~\cite{lu20253dgs} dataset, which focuses on tabletop scenes with complex object changes, including object removal, insertion, and movement in cluttered environments. 

\paragraph{Baselines and Metrics}  We compare our method with 2D-based approaches, \ie, MTP~\cite{wang2024mtp}, and 3D-based methods 3DGS-CD~\cite{lu20253dgs}, MV3DCD~\cite{Galappaththige_2025_CVPR}, and the change detection module of CLSplat~\cite{ackermann2025clsplats}. In addition, we evaluate two baselines based on pixel difference (Pixel-Diff) and feature difference (Feature-Diff).
Following the evaluation protocols in C-NeRF~\cite{huang2023c} and 3DGS-CD~\cite{lu20253dgs}, we report Precision, Recall, F1-score, and IoU as our quantitative metrics.

\subsection{Evaluation Results}
\label{sec:EvaRes}

% Evaluation Results: change detection (\datasetname dataset) + reconstruction

\datasetname Results in  \cref{tab:blender} show that our method consistently achieves the highest Precision and IoU across all scenes. 2D pixel and feature differences yield stable but worse results due to noise from Gaussian splatting artifacts. MV3DCD’s Gaussian-wise change representation often leads to fragmented detections lacking object integrity. Furthermore, 3DGS-CD's performance is highly sensitive to its object matching and pose estimation steps, especially in challenging scenarios. For instance, it fails in the Bedroom scene, where limited viewpoints lead to critical matching errors, and struggles to distinguish visually similar books in the Bookcase scene, causing a significant drop in IoU. In contrast, our approach uses 3D difference aggregation and segmentation validation to effectively suppress such noise and recover complete, accurate change masks.

\begin{table}[ht]
\centering
\caption{\textbf{Controlled Evaluation on the Livingroom scene.} We report F1/IoU across change types and number of changed objects.}
\label{tab:livingroom_eval}
\resizebox{\linewidth}{!}{
\begin{tabular}{l *{3}{cc}}
\toprule
& \multicolumn{2}{c}{\# Obj = 1} & \multicolumn{2}{c}{\# Obj = 2} & \multicolumn{2}{c}{\# Obj = 4} \\
\cmidrule(lr){2-3}\cmidrule(lr){4-5}\cmidrule(lr){6-7}
\textbf{Change Type} & \textbf{F1} & \textbf{IoU} & \textbf{F1} & \textbf{IoU} & \textbf{F1} & \textbf{IoU} \\
\midrule
In/Out       & 0.953 & 0.911 & 0.770 & 0.742 & 0.764 & 0.732 \\
Translation  & 0.985 & 0.970 & 0.613 & 0.498 & 0.585 & 0.472 \\
Rotation     & 0.982 & 0.964 & 0.690 & 0.613 & 0.383 & 0.253 \\
Mixed        &   --  &   --  & 0.604 & 0.582 & 0.408 & 0.260 \\
\bottomrule
\end{tabular}
}
\end{table}

\paragraph{Controlled Experiments} We evaluate our method under varying change types and object counts. \cref{tab:livingroom_eval} shows that performance remains strong for simple scenarios (\eg, single-object cases) across all change types.
However, as the number of changed objects increases, performance degrades, especially for Rotation and Translation. Mixed changes are the most challenging overall, yielding lower scores and compounded difficulty under heterogeneous transformations. This also validates the utility of our dataset in examining algorithm robustness in more complex cases.

% \begin{table}[ht]
% \centering
% \caption{\textbf{Evaluation of our method on the Livingroom scene (\datasetname).} The first column lists the change types present in the scene. The second column shows the number of changed objects.}
% \label{tab:detailed_results}
% \begin{tabular}{lccc}
% \toprule
% \textbf{Change Type} & \textbf{\# Obj} & \textbf{F1} & \textbf{IoU} \\
% \midrule
% In and Out  & 1 & 0.953 & 0.911 \\
% In and Out  & 2 & 0.770 & 0.742 \\
% In and Out  & 4 & 0.764 & 0.732 \\
% Translation & 1 & 0.985 & 0.970 \\
% Translation & 2 & 0.613 & 0.498 \\
% Translation & 4 & 0.585 & 0.472 \\
% Rotation    & 1 & 0.982 & 0.964 \\
% Rotation    & 2 & 0.690 & 0.613 \\
% Rotation    & 4 & 0.383 & 0.253 \\
% Mixed       & 2 & 0.390 & 0.307 \\
% Mixed       & 4 & 0.408 & 0.260 \\
% \bottomrule
% \end{tabular}
% \label{tab:Control_dataset}
% \end{table}

% the real-world dataset (3DGS-CD dataset) -> supp
\paragraph{3DGS-CD Dataset} Results on the 3DGS-CD dataset (\cref{tab:3dgs_results}) show our method achieves significantly higher F1 and IoU scores than prior approaches. While MTP attains high precision, its lower recall limits performance. The 3DGS-CD method localizes changes accurately but suffers from instability due to reliance on 2D detection and object association steps, especially in cluttered scenes like Mustard and Bench. In contrast, our approach delivers more consistent results by leveraging 3D difference voting and validation to reduce errors from 2D difference detection.

\begin{table}
  \centering
  \small
  \caption{\textbf{Quantitative change detection results on the 3DGS-CD dataset.} Our method consistently achieves the best performance in terms of F1 score and IoU.}
  \begin{tabular}{l l c c c c}
    \toprule
    \textbf{Scene} & \textbf{Method} & \textbf{Precision} & \textbf{Recall} & \textbf{F1} & \textbf{IoU} \\
    \midrule
    \multirow{3}{*}{Mustard} 
        & MTP     & \cellcolor{gbest}0.949 & \cellcolor{gsecond}0.231 & \cellcolor{gsecond}0.371 & \cellcolor{gsecond}0.228 \\
        & 3DGS-CD & 0.315 & 0.104 & 0.155 & 0.085 \\
        & Ours    & \cellcolor{gsecond}0.794 & \cellcolor{gbest}0.573 & \cellcolor{gbest}0.583 & \cellcolor{gbest}0.507 \\
    \addlinespace
    
    \multirow{3}{*}{Desk} 
        & MTP      & 0.957 & 0.344 & 0.506 & 0.339 \\
        & 3DGS-CD  & \cellcolor{gsecond}0.967 & \cellcolor{gsecond}0.961 & \cellcolor{gsecond}0.964 & \cellcolor{gsecond}0.930 \\
        & Ours     & \cellcolor{gbest}0.995 & \cellcolor{gbest}0.968 & \cellcolor{gbest}0.981 & \cellcolor{gbest}0.964 \\
    \addlinespace
    
    \multirow{3}{*}{Swap} 
        & MTP      & 0.942 & 0.246 & 0.390 & 0.243 \\
        & 3DGS-CD  & \cellcolor{gsecond}0.983 & \cellcolor{gsecond}0.989 & \cellcolor{gsecond}0.986 & \cellcolor{gsecond}0.973 \\
        & Ours     & \cellcolor{gbest}0.998 & \cellcolor{gbest}0.992 & \cellcolor{gbest}0.995 & \cellcolor{gbest}0.990 \\
    \addlinespace
    
    \multirow{3}{*}{Bench} 
        & MTP      & \cellcolor{gsecond}0.902 & \cellcolor{gbest}0.887 & \cellcolor{gsecond}0.895 & \cellcolor{gsecond}0.809 \\
        & 3DGS-CD  & 0.851 & 0.796 & 0.817 & 0.691 \\
        & Ours     & \cellcolor{gbest}0.995 & \cellcolor{gsecond}0.867 & \cellcolor{gbest}0.915 & \cellcolor{gbest}0.863 \\
    \addlinespace
    
    \multirow{3}{*}{Sill} 
        & MTP      & 0.483 & 0.308 & 0.376 & 0.232 \\
        & 3DGS-CD  & \cellcolor{gsecond}0.981 & \cellcolor{gbest}0.974 & \cellcolor{gsecond}0.977 & \cellcolor{gsecond}0.956 \\
        & Ours     & \cellcolor{gbest}0.998 & \cellcolor{gsecond}0.972 & \cellcolor{gbest}0.982 & \cellcolor{gbest}0.970 \\
    \midrule
    
    \multirow{3}{*}{Average} 
        & MTP      & \cellcolor{gsecond}0.846 & 0.403 & 0.508 & 0.370 \\
        & 3DGS-CD  & 0.819 & \cellcolor{gsecond}0.765 & \cellcolor{gsecond}0.780 & \cellcolor{gsecond}0.727 \\
        & Ours     & \cellcolor{gbest}0.956 & \cellcolor{gbest}0.874 & \cellcolor{gbest}0.891 & \cellcolor{gbest}0.859 \\
    \bottomrule
  \end{tabular}
  \label{tab:3dgs_results}
\end{table}
% For the 3DGS-CD dataset, the results are presented in \cref{tab:3dgs_results}. Our method achieves substantially higher F1 scores and IoU compared to previous approaches. MTP~\cite{wang2024mtp}, a state-of-the-art 2D change detection method, is a foundation model pre-trained on remote sensing datasets for multiple tasks. It demonstrates high precision but lower recall, indicating a tendency to miss certain changed regions, thereby leading to reduced F1 scores and IoU.   

% The 3DGS-CD method can localize changed regions with high accuracy. However, its performance heavily depends on the 2D change detection step and the change-aware object association step, resulting in high score variance. For instance, in the Mustard dataset, the changed bottle was placed against a complex background containing many other bottles and boxes, which confused the EfficientSAM-based 2D difference detector and led to incorrect predictions. In the Bench dataset, a large number of changed objects biased the change-object association and pose estimation processes, thereby reducing performance.   

% In contrast, our method exhibits more stable performance. While it is still affected by the 2D difference detection errors in the Mustard dataset, the subsequent 3D difference voting and validation steps help mitigate this issue, resulting in consistently higher scores.

\begin{table}[h]
\caption{\textbf{Quantitative reconstruction results on novel views.} Change-centric crops refer to test-view images cropped using bounding boxes that tightly enclose the ground-truth change regions, while Full scenes correspond to the entire test-view images.}
% Our method achieves the best overall performance on full scenes and competitive results on change-centric crops.
\centering
\small
\begin{tabular}{lccc}
\toprule
Method & \makecell{PSNR$\uparrow$} & \makecell{SSIM$\uparrow$} & \makecell{LPIPS$\downarrow$} \\
\midrule
\rowcolor{gray!30} \multicolumn{4}{l}{\textbf{\textit{Change-centric crops}}} \\ 
Post-on-Pre & \cellcolor{gbest}21.996 & \cellcolor{gsecond}0.606 & \cellcolor{gbest}0.333 \\
Mixed & 18.251 & 0.576 & 0.550 \\
GaussianEditor & 17.356 & 0.554 & 0.585 \\
Change Mask & 13.318 & 0.445 & 0.628 \\
Ours & \cellcolor{gsecond}21.837 & \cellcolor{gbest}0.608 & \cellcolor{gsecond}0.356 \\
\midrule
\rowcolor{gray!30} \multicolumn{4}{l}{\textbf{\textit{Full scenes}}} \\ 
Post-on-Pre & 27.680 & 0.900 & 0.173 \\
Mixed & \cellcolor{gsecond}27.982 & \cellcolor{gbest}0.944 & \cellcolor{gsecond}0.117 \\
GaussianEditor & 22.348 & 0.867 & 0.252 \\
Change Mask & 16.114 & 0.730 & 0.479 \\
Ours & \cellcolor{gbest}30.304 & \cellcolor{gsecond}0.939 & \cellcolor{gbest}0.116 \\
\bottomrule
\end{tabular}
\label{tab:recon_novel}
\end{table}

% \begin{table}[h]
% \centering
% \small
% \caption{\textbf{Quantitative reconstruction results.} Scenes are rendered from both post-change and novel viewpoints. }
% \begin{tabular}{lccc}
% \toprule
% % Method & Metric A & Metric B & Metric C \\
% Method & \makecell{PSNR$\uparrow$} & \makecell{LPIPS$\downarrow$} & \makecell{SSIM$\uparrow$} \\
% \midrule
% \rowcolor{gray!30} \multicolumn{4}{l}{\textbf{\textit{Post-change views}}} \\ 
% Post-on-Pre        & \cellcolor{gbest}34.496 & \cellcolor{gbest}0.07  & \cellcolor{gbest}0.945 \\
% Mixed     & 26.665 & 0.231 & 0.855 \\
% GaussianEditor     & 20.664 & 0.446 & 0.722 \\
% Change Mask      & 25.708 & 0.262 & 0.826 \\
% Ours              & \cellcolor{gsecond}33.095 & \cellcolor{gsecond}0.083 & \cellcolor{gsecond}0.935 \\
% \midrule
% \rowcolor{gray!30} \multicolumn{4}{l}{\textbf{\textit{Novel views}}} \\ 
% Post-on-Pre     & 26.954 & 0.166 & 0.888 \\
% Mixed  & \cellcolor{gsecond}30.345 & \cellcolor{gsecond}0.095 & \cellcolor{gsecond}0.936 \\
% GaussianEditor  & 19.338 & 0.366 & 0.765 \\
% Change Mask    & 25.324 & 0.202 & 0.874 \\
% Ours            & \cellcolor{gbest}32.115 & \cellcolor{gbest}0.087 & \cellcolor{gbest}0.938 \\
% \bottomrule
% \end{tabular}
% \label{tab:recon_novel}
% \end{table}

% ... achieves the best performance on post-change views but
\paragraph{Continual Scene Reconstruction} We further evaluate the image quality of our continual reconstruction results. As reported in \cref{tab:recon_novel}, our method achieves the highest rendering quality on full-scene novel views and delivers the second-best performance on change-centric crops, demonstrating strong overall robustness across both global and change-focused regions. Directly fine-tuning the pre-change Gaussians $\mathcal{G}{pre}$ with post-change images $\mathcal{I}{post}$ suffers from severe view-dependent artifacts in full scenes. Training with mixed pre- and post-change images tends to retain pre-change object states because post-change views are far fewer, leading to lower performance on change-centric crops. Qualitative comparisons are provided in \cref{fig:Recon} and \cref{fig:ReconPost}. As illustrated in \cref{fig:Recon}, GaussianEditor~\cite{GaussianEditor} often fails to precisely identify the correct new location or state of the changed objects.

\subsection{Ablation Study}
\label{sec:Ablation}
\cref{tab:ablation} presents the results of our ablation study, averaged over the \datasetname dataset. The first five rows correspond to vanilla 2D change detection methods commonly adopted in prior works~\cite{fu2025gslts3dgaussiansplattingbased, ackermann2025clsplats, lu20253dgs, Galappaththige_2025_CVPR}, indicating that feature difference is more robust than pixel- and SSIM-based difference, while combining them via multiplication does not yield significant improvement. Compared to 2D difference methods, the 3D difference approaches consistently achieve better performance, demonstrating the effectiveness of multi-view voting in suppressing noise. Our full method, which integrates 2D difference, 3D difference, and segmentation validation, attains the highest accuracy overall.
\begin{table}[ht]
\small
\centering
\caption{\textbf{Ablation study.} We evaluate pixel-, feature-, and SSIM-based differences, as well as their combinations.
% , \ie, pixel+feature~\cite{fu2025gslts3dgaussiansplattingbased} and SSIM+feature~\cite{Galappaththige_2025_CVPR}). 
They are compared against 3D difference and our full method. 
% For 2D methods, input images are pre-aligned to the test views.
}
\label{tab:ablation}
\begin{tabular}{lcccc}
\toprule
\textbf{Method}          & \textbf{Precision} & \textbf{Recall} & \textbf{F1} & \textbf{IoU} \\
\midrule
\rowcolor{gray!30} \multicolumn{5}{l}{\textbf{\textit{2D difference}}} \\ 
Pixel-Diff                   & 0.303              & 0.458           & 0.318       & 0.198        \\
Feature-Diff                  & 0.436              & 0.538           & \cellcolor{gthird}0.450       &\cellcolor{gthird} 0.343        \\
SSIM-Diff                 & 0.316              & \cellcolor{gsecond}0.637           & 0.394       & 0.256        \\
Pixel+Feature         &\cellcolor{gthird} 0.519              & 0.241           & 0.308       & 0.199        \\
SSIM+Feature        & 0.481              & 0.355           & 0.388       & 0.271        \\
\midrule
\rowcolor{gray!30} \multicolumn{5}{l}{\textbf{\textit{3D difference}}} \\ 
3D-Diff                    & \cellcolor{gsecond}0.522              & \cellcolor{gthird}0.617           & \cellcolor{gsecond}0.487       &\cellcolor{gsecond} 0.366        \\
Ours (Full)                & \cellcolor{gbest} 0.836              & \cellcolor{gbest}0.680           & \cellcolor{gbest}0.724       &\cellcolor{gbest} 0.650        \\
\bottomrule
\end{tabular}
\end{table}

% \begin{table}[h]
% \centering
% \caption{\textbf{Efficiency evaluation on Livingroom scene (\modelname).} (Processing Time in Seconds)}
% \label{tab:efficiency}
% \begin{tabular}{lcccccc}
% \toprule
% Method & PixDif & FeatDif & CL-Splat & MV3DCD & 3GDS-CD & Ours \\
% \midrule
% Time (s) & 0.6 & 28.61 & 36.25 & 434.68 & 58.72 & 31.03 \\
% \bottomrule
% \end{tabular}
% \end{table}

% caption
% Pixel Feature(EfficientSAM)  SSIM  Pixel*Feature Feature*SSIM Pixel*SSIM 
% 3DDifference-trace 3DDiff-MV3DCD 3DDif-CLSplat Ours-full

\section{Conclusion}

% We propose \modelname, a multi-view voting-and-validation-based 3D change detection and reconstruction framework designed for complex and large-scale 3D scenes. By integrating 2D signed-distance differencing with multi-view aggregation and pruning, our method generates accurate and consistent change masks, enabling localized updates to dynamic 3D scenes. Extensive experiments, together with the proposed dataset \datasetname, demonstrate that \modelname outperforms state-of-the-art methods in both accuracy and efficiency, yielding high-fidelity reconstructions with fewer artifacts. Future work may address current limitations, including handling a greater number of changed objects within a scene and mitigating the effects of varying lighting conditions and shadows. Furthermore, exploring how to effectively model non-rigid deformations or significant topological changes within the 3D Gaussian framework remains a key challenge.

We propose \modelname, a multi-view voting and validation framework for 3D change detection and reconstruction in complex and large-scale 3D scenes. Our method generates accurate and consistent change masks, enabling localized updates to dynamic 3D scenes. Extensive experiments, together with the proposed dataset \datasetname, demonstrate that \modelname outperforms state-of-the-art methods in both accuracy and efficiency. Future work may address current limitations, including handling a greater number of changed objects within a scene and mitigating the effects of varying lighting conditions and shadows. Furthermore, exploring how to effectively model non-rigid deformations or significant topological changes within the 3D Gaussian framework remains a key challenge.

\clearpage
{
    \small
    \bibliographystyle{ieeenat_fullname}
    \bibliography{reference}

@String(CVPR= {IEEE Conf. Comput. Vis. Pattern Recog.})

@String(ICCV= {Int. Conf. Comput. Vis.})

@String(ECCV= {Eur. Conf. Comput. Vis.})

@String(TOG= {ACM Trans. Graph.})

@String(AAAI = {AAAI})

@String(CVPR  = {CVPR})

@String(ICCV  = {ICCV})

@String(ECCV  = {ECCV})

@String(TOG   = {ACM TOG})

@inproceedings{yariv2021volume,
  title={Volume rendering of neural implicit surfaces},
  author={Yariv, Lior and Gu, Jiatao and Kasten, Yoni and Lipman, Yaron},
  booktitle={Thirty-Fifth Conference on Neural Information Processing Systems},
  year={2021}
}

@article{wang2021neus,
  title={NeuS: Learning Neural Implicit Surfaces by Volume Rendering for Multi-view Reconstruction},
  author={Wang, Peng and Liu, Lingjie and Liu, Yuan and Theobalt, Christian and Komura, Taku and Wang, Wenping},
  journal={arXiv preprint arXiv:2106.10689},
  year={2021}
}

@Article{kerbl3Dgaussians,
      title        = {3D Gaussian Splatting for Real-Time Radiance Field Rendering},
      author       = {Kerbl, Bernhard and Kopanas, Georgios and Leimk{\"u}hler, Thomas and Drettakis, George},
      journal      = {ACM Transactions on Graphics},
      number       = {4},
      volume       = {42},
      month        = {July},
      year         = {2023},
      url          = {https://repo-sam.inria.fr/fungraph/3d-gaussian-splatting/}
}

@inproceedings{mildenhall2020nerf,
  title={NeRF: Representing Scenes as Neural Radiance Fields for View Synthesis},
  author={Ben Mildenhall and Pratul P. Srinivasan and Matthew Tancik and Jonathan T. Barron and Ravi Ramamoorthi and Ren Ng},
  year={2020},
  booktitle={ECCV},
}

@article{dang2024beyond,
  title={Beyond appearance: Multi-frame spatio-temporal context memory networks for efficient and robust video object segmentation},
  author={Dang, Jisheng and Zheng, Huicheng and Xu, Xiaohao and Wang, Longguang and Guo, Yulan},
  journal={IEEE Transactions on Image Processing},
  year={2024},
  publisher={IEEE}
}

@article{lv2024weakly,
  title={Weakly-supervised contrastive learning for unsupervised object discovery},
  author={Lv, Yunqiu and Zhang, Jing and Barnes, Nick and Dai, Yuchao},
  journal={IEEE Transactions on Image Processing},
  volume={33},
  pages={2689--2702},
  year={2024},
  publisher={IEEE}
}

@article{mackiewicz1993principal,
  title={Principal components analysis (PCA)},
  author={Ma{\'c}kiewicz, Andrzej and Ratajczak, Waldemar},
  journal={Computers \& Geosciences},
  volume={19},
  number={3},
  pages={303--342},
  year={1993},
  publisher={Elsevier}
}

@inproceedings{schoenberger2016sfm,
    author={Sch\"{o}nberger, Johannes Lutz and Frahm, Jan-Michael},
    booktitle={Conference on Computer Vision and Pattern Recognition (CVPR)},
    title={Structure-from-Motion Revisited},
    year={2016},
}

@inproceedings{schoenberger2016mvs,
    title={Pixelwise View Selection for Unstructured Multi-View Stereo},
    author={Sch\"{o}nberger, Johannes Lutz and Zheng, Enliang and Pollefeys, Marc and Frahm, Jan-Michael},
    booktitle={European Conference on Computer Vision (ECCV)},
    year={2016},
}

@article{xiong2023efficientsam,
  title={EfficientSAM: Leveraged Masked Image Pretraining for Efficient Segment Anything},
  author={Yunyang Xiong and Bala Varadarajan and Lemeng Wu and Xiaoyu Xiang and Fanyi Xiao and Chenchen Zhu and Xiaoliang Dai and Dilin Wang and Fei Sun and Forrest Iandola and Raghuraman Krishnamoorthi and Vikas Chandra},
  journal={arXiv:2312.00863},
  year={2023}
}

@inproceedings{caron2021emerging,
  title={Emerging Properties in Self-Supervised Vision Transformers},
  author={Caron, Mathilde and Touvron, Hugo and Misra, Ishan and J\'egou, Herv\'e  and Mairal, Julien and Bojanowski, Piotr and Joulin, Armand},
  booktitle={Proceedings of the International Conference on Computer Vision (ICCV)},
  year={2021}
}

@article{bovolo2006theoretical,
  title={A theoretical framework for unsupervised change detection based on change vector analysis in the polar domain},
  author={Bovolo, Francesca and Bruzzone, Lorenzo},
  journal={IEEE Transactions on Geoscience and Remote Sensing},
  volume={45},
  number={1},
  pages={218--236},
  year={2006},
  publisher={IEEE}
}

@article{celik2009unsupervised,
  title={Unsupervised change detection in satellite images using principal component analysis and $ k $-means clustering},
  author={Celik, Turgay},
  journal={IEEE geoscience and remote sensing letters},
  volume={6},
  number={4},
  pages={772--776},
  year={2009},
  publisher={IEEE}
}

@article{luppino2019unsupervised,
  title={Unsupervised image regression for heterogeneous change detection},
  author={Luppino, Luigi T and Bianchi, Filippo M and Moser, Gabriele and Anfinsen, Stian N},
  journal={arXiv preprint arXiv:1909.05948},
  year={2019}
}

@article{li2025dynamicearth,
  title={DynamicEarth: How Far are We from Open-Vocabulary Change Detection?},
  author={Li, Kaiyu and Cao, Xiangyong and Deng, Yupeng and Pang, Chao and Xin, Zepeng and Meng, Deyu and Wang, Zhi},
  journal={arXiv preprint arXiv:2501.12931},
  year={2025}
}

@article{fang2022changer,
  title={Changer: Feature Interaction is What You Need for Change Detection},
  author={Fang, Sheng and Li, Kaiyu and Li, Zhe},
  journal={arXiv preprint arXiv:2209.08290},
  year={2022}
}

@inproceedings{bandara2022transformer,
  title={A transformer-based siamese network for change detection},
  author={Bandara, Wele Gedara Chaminda and Patel, Vishal M},
  booktitle={IGARSS 2022-2022 IEEE International Geoscience and Remote Sensing Symposium},
  pages={207--210},
  year={2022},
  organization={IEEE}
}

@Article{chen2021a,
    title={Remote Sensing Image Change Detection with Transformers},
    author={Hao Chen, Zipeng Qi and Zhenwei Shi},
    year={2021},
    journal={IEEE Transactions on Geoscience and Remote Sensing},
    volume={},
    number={},
    pages={1-14},
    doi={10.1109/TGRS.2021.3095166}
}

@article{dong2024changeclip,
  title={ChangeCLIP: Remote sensing change detection with multimodal vision-language representation learning},
  author={Dong, Sijun and Wang, Libo and Du, Bo and Meng, Xiaoliang},
  journal={ISPRS Journal of Photogrammetry and Remote Sensing},
  volume={208},
  pages={53--69},
  year={2024},
  publisher={Elsevier}
}

@article{wang2024mtp,
  title={Mtp: Advancing remote sensing foundation model via multi-task pretraining},
  author={Wang, Di and Zhang, Jing and Xu, Minqiang and Liu, Lin and Wang, Dongsheng and Gao, Erzhong and Han, Chengxi and Guo, Haonan and Du, Bo and Tao, Dacheng and others},
  journal={IEEE Journal of Selected Topics in Applied Earth Observations and Remote Sensing},
  year={2024},
  publisher={IEEE}
}

@article{sakurada2020weakly,
  title={Weakly Supervised Silhouette-based Semantic Scene Change Detection},
  author={Sakurada, Ken and Shibuya, Mikiya and Wang Weimin},
  journal={Proceedings of the IEEE International Conference on Robotics and Automation (ICRA)},
  year={2020}
}

@inproceedings{martinson2024meaningful,
  title={Meaningful Change Detection in Indoor Environments Using CLIP Models and NeRF-Based Image Synthesis},
  author={Martinson, Eric and Lauren, Paula},
  booktitle={2024 21st International Conference on Ubiquitous Robots (UR)},
  pages={603--610},
  year={2024},
  organization={IEEE}
}

@article{huang2023c,
  title={C-NERF: Representing Scene Changes as Directional Consistency Difference-based NeRF},
  author={Huang, Rui and Jiang, Binbin and Zhao, Qingyi and Wang, William and Zhang, Yuxiang and Guo, Qing},
  journal={arXiv preprint arXiv:2312.02751},
  year={2023}
}

@InProceedings{Galappaththige_2025_CVPR,
title = {Multi-View Pose-Agnostic Change Localization with Zero Labels},
author = {Galappaththige, Chamuditha Jayanga and Lai, Jason and Windrim, Lloyd and Dansereau, Donald and Sunderhauf, Niko and Miller, Dimity},
booktitle = {Proceedings of the Computer Vision and Pattern Recognition Conference (CVPR)},
month = {June},
year = {2025},
pages = {11600-11610}
}

@article{lu20253dgs,
  title={3DGS-CD: 3D Gaussian Splatting-Based Change Detection for Physical Object Rearrangement},
  author={Lu, Ziqi and Ye, Jianbo and Leonard, John},
  journal={IEEE Robotics and Automation Letters},
  year={2025},
  publisher={IEEE}
}

@article{de2021continual,
  title={A continual learning survey: Defying forgetting in classification tasks},
  author={De Lange, Matthias and Aljundi, Rahaf and Masana, Marc and Parisot, Sarah and Jia, Xu and Leonardis, Ale{\v{s}} and Slabaugh, Gregory and Tuytelaars, Tinne},
  journal={IEEE transactions on pattern analysis and machine intelligence},
  volume={44},
  number={7},
  pages={3366--3385},
  year={2021},
  publisher={IEEE}
}

@inproceedings{wang2023sparsenerf,
  title={Sparsenerf: Distilling depth ranking for few-shot novel view synthesis},
  author={Wang, Guangcong and Chen, Zhaoxi and Loy, Chen Change and Liu, Ziwei},
  booktitle={Proceedings of the IEEE/CVF international conference on computer vision},
  pages={9065--9076},
  year={2023}
}

@inproceedings{palazzolo2019refusion,
  title={ReFusion: 3D reconstruction in dynamic environments for RGB-D cameras exploiting residuals},
  author={Palazzolo, Emanuele and Behley, Jens and Lottes, Philipp and Giguere, Philippe and Stachniss, Cyrill},
  booktitle={2019 IEEE/RSJ International Conference on Intelligent Robots and Systems (IROS)},
  pages={7855--7862},
  year={2019},
  organization={IEEE}
}

@inproceedings{li2024learn,
  title={Learn to Memorize and to Forget: A Continual Learning Perspective of Dynamic SLAM},
  author={Li, Baicheng and Yan, Zike and Wu, Dong and Jiang, Hanqing and Zha, Hongbin},
  booktitle={European Conference on Computer Vision},
  pages={41--57},
  year={2024},
  organization={Springer}
}

@misc{fu2025gslts3dgaussiansplattingbased,
      title={GS-LTS: 3D Gaussian Splatting-Based Adaptive Modeling for Long-Term Service Robots}, 
      author={Bin Fu and Jialin Li and Bin Zhang and Ruiping Wang and Xilin Chen},
      year={2025},
      eprint={2503.17733},
      archivePrefix={arXiv},
      primaryClass={cs.RO},
      url={https://arxiv.org/abs/2503.17733}, 
}

@article{kulhanek2024wildgaussians,
  title={WildGaussians: 3D Gaussian Splatting in the Wild},
  author={Kulhanek, Jonas and Peng, Songyou and Kukelova, Zuzana and Pollefeys, Marc and Sattler, Torsten},
  journal={NeurIPS},
  year={2024}
}

@article{ackermann2025clsplats,
  title     = {CL-Splats: Continual Learning Gaussian Splatting with Local Optimization},
  author    = {Ackermann, Jan and Kulhanek, Jonas and Cai, Shengqu and Haofei, Xu and Pollefeys, Marc and Wetzstein, Gordon and Guibas, Leonidas and Peng, Songyou},
  journal   = {ICCV},
  year      = {2025},
}

@inproceedings{cai2023clnerf,
  title={Clnerf: Continual learning meets nerf},
  author={Cai, Zhipeng and M{\"u}ller, Matthias},
  booktitle={Proceedings of the IEEE/CVF International Conference on Computer Vision},
  pages={23185--23194},
  year={2023}
}

@article{zhuang2024tip,
  title={Tip-editor: An accurate 3d editor following both text-prompts and image-prompts},
  author={Zhuang, Jingyu and Kang, Di and Cao, Yan-Pei and Li, Guanbin and Lin, Liang and Shan, Ying},
  journal={ACM Transactions on Graphics (TOG)},
  volume={43},
  number={4},
  pages={1--12},
  year={2024},
  publisher={ACM New York, NY, USA}
}

@inproceedings{GaussianEditor,
  title = {GaussianEditor: Editing 3D Gaussians Delicately with Text Instructions},
  author = {Fang, Jiemin and Wang, Junjie and Zhang, Xiaopeng and Xie, Lingxi and Tian, Qi},
  year = {2024},
  booktitle = {CVPR}
}

@article{palandra2024gsedit,
  title={Gsedit: Efficient text-guided editing of 3d objects via gaussian splatting},
  author={Palandra, Francesco and Sanchietti, Andrea and Baieri, Daniele and Rodola, Emanuele},
  journal={arXiv preprint arXiv:2403.05154},
  year={2024}
}

@inproceedings{wu2024gaussctrl,
  title={Gaussctrl: Multi-view consistent text-driven 3d gaussian splatting editing},
  author={Wu, Jing and Bian, Jia-Wang and Li, Xinghui and Wang, Guangrun and Reid, Ian and Torr, Philip and Prisacariu, Victor Adrian},
  booktitle={European Conference on Computer Vision},
  pages={55--71},
  year={2024},
  organization={Springer}
}

@article{jaganathan2024ice,
  title={Ice-g: Image conditional editing of 3d gaussian splats},
  author={Jaganathan, Vishnu and Huang, Hannah Hanyun and Irshad, Muhammad Zubair and Jampani, Varun and Raj, Amit and Kira, Zsolt},
  journal={arXiv preprint arXiv:2406.08488},
  year={2024}
}

@article{yan20243dsceneeditor, 
  title={3dsceneeditor: Controllable 3d scene editing with gaussian splatting},
  author={Yan, Ziyang and Li, Lei and Shao, Yihua and Chen, Siyu and Wu, Zongkai and Hwang, Jenq-Neng and Zhao, Hao and Remondino, Fabio},
  journal={arXiv preprint arXiv:2412.01583},
  year={2024}
}

@inproceedings{gaussian_grouping,
    title={Gaussian Grouping: Segment and Edit Anything in 3D Scenes},
    author={Ye, Mingqiao and Danelljan, Martin and Yu, Fisher and Ke, Lei},
    booktitle={ECCV},
    year={2024}
}

@inproceedings{cen2025segment,
  title={Segment any 3d gaussians},
  author={Cen, Jiazhong and Fang, Jiemin and Yang, Chen and Xie, Lingxi and Zhang, Xiaopeng and Shen, Wei and Tian, Qi},
  booktitle={Proceedings of the AAAI Conference on Artificial Intelligence},
  volume={39},
  number={2},
  pages={1971--1979},
  year={2025}
}

@misc{guo2024semantic, 
    title={Semantic Gaussians: Open-Vocabulary Scene Understanding with 3D Gaussian Splatting}, 
    author={Jun Guo and Xiaojian Ma and Yue Fan and Huaping Liu and Qing Li},
    year={2024},
    eprint={2403.15624},
    archivePrefix={arXiv},
    primaryClass={cs.CV}
  }

@article{gu2024egolifter,
  author    = {Gu, Qiao and Lv, Zhaoyang and Frost, Duncan and Green, Simon and Straub, Julian and Sweeney, Chris},
  title     = {EgoLifter: Open-world 3D Segmentation for Egocentric Perception},
  journal   = {arXiv preprint arXiv:2403.18118},
  year      = {2024},
}

@article{wang2021clip,
  title={CLIP-NeRF: Text-and-Image Driven Manipulation of Neural Radiance Fields},
  author={Wang, Can and Chai, Menglei and He, Mingming and Chen, Dongdong and Liao, Jing},
  journal={arXiv preprint arXiv:2112.05139},
  year={2021}
}

@inproceedings{bao2023sine,
    title={SINE: Semantic-driven Image-based NeRF Editing with Prior-guided Editing Field},
    author={Bao, Chong and Zhang, Yinda and Yang, Bangbang and Fan, Tianxing and Yang, Zesong and Bao, Hujun and Zhang, Guofeng and Cui, Zhaopeng},
    booktitle={The IEEE/CVF Computer Vision and Pattern Recognition Conference (CVPR)},
    year={2023}
}

@inproceedings{zhang2021stnerf,
   title={Editable Free-Viewpoint Video using a Layered Neural Representation},
   author={Jiakai, Zhang and Xinhang, Liu and Xinyi, Ye and Fuqiang, Zhao and Yanshun, Zhang and Minye, Wu and Yingliang, Zhang and Lan, Xu and Jingyi, Yu},
 year={2021},
 booktitle={ACM SIGGRAPH},
    }

@misc{blender,
  author={Blender Online Community},
  title={Blender - a 3D modelling and rendering package.},
  howpublished={Blender Foundation, Stichting Blender Foundation, Amsterdam},
  year={2018},
}

@article{jain2024gaussiancut,
  title={Gaussiancut: Interactive segmentation via graph cut for 3d gaussian splatting},
  author={Jain, Umangi and Mirzaei, Ashkan and Gilitschenski, Igor},
  journal={Advances in Neural Information Processing Systems},
  volume={37},
  pages={89184--89212},
  year={2024}
}

@inproceedings{taneja2011image,
  title={Image based detection of geometric changes in urban environments},
  author={Taneja, Aparna and Ballan, Luca and Pollefeys, Marc},
  booktitle={2011 international conference on computer vision},
  pages={2336--2343},
  year={2011},
  organization={IEEE}
}

@inproceedings{taneja2013city,
  title={City-scale change detection in cadastral 3d models using images},
  author={Taneja, Aparna and Ballan, Luca and Pollefeys, Marc},
  booktitle={Proceedings of the IEEE Conference on computer Vision and Pattern Recognition},
  pages={113--120},
  year={2013}
}

@inproceedings{fehr2017tsdf,
  title={TSDF-based change detection for consistent long-term dense reconstruction and dynamic object discovery},
  author={Fehr, Marius and Furrer, Fadri and Dryanovski, Ivan and Sturm, J{\"u}rgen and Gilitschenski, Igor and Siegwart, Roland and Cadena, Cesar},
  booktitle={2017 IEEE International Conference on Robotics and automation (ICRA)},
  pages={5237--5244},
  year={2017},
  organization={IEEE}
}

@inproceedings{palazzolo2017change,
  title={Change detection in 3d models based on camera images},
  author={Palazzolo, Emanuele and Stachniss, Cyrill},
  booktitle={9th Workshop on Planning, Perception and Navigation for Intelligent Vehicles at the IEEE/RSJ Int. Conf. on Intelligent Robots and Systems (IROS)},
  volume={3},
  pages={14},
  year={2017}
}

@inproceedings{palazzolo2018fast,
  title={Fast image-based geometric change detection given a 3d model},
  author={Palazzolo, Emanuele and Stachniss, Cyrill},
  booktitle={2018 IEEE International Conference on Robotics and Automation (ICRA)},
  pages={6308--6315},
  year={2018},
  organization={IEEE}
}

@inproceedings{langer2020robust,
  title={Robust and efficient object change detection by combining global semantic information and local geometric verification},
  author={Langer, Edith and Patten, Timothy and Vincze, Markus},
  booktitle={2020 IEEE/RSJ International Conference on Intelligent Robots and Systems (IROS)},
  pages={8453--8460},
  year={2020},
  organization={IEEE}
}

@inproceedings{adam2022objects,
  title={Objects can move: 3d change detection by geometric transformation consistency},
  author={Adam, Aikaterini and Sattler, Torsten and Karantzalos, Konstantinos and Pajdla, Tomas},
  booktitle={European Conference on Computer Vision},
  pages={108--124},
  year={2022},
  organization={Springer}
}

@inproceedings{ni2025dprecon,
  title={Decompositional Neural Scene Reconstruction with Generative Diffusion Prior},
  author={Ni, Junfeng and Liu, Yu and Lu, Ruijie and Zhou, Zirui and Zhu, Song-Chun and Chen, Yixin and Huang, Siyuan},
  booktitle=CVPR,
  year={2025}
}

@inproceedings{shen2025trace3d,
  title={Trace3D: Consistent Segmentation Lifting via Gaussian Instance Tracing},
  author={Shen, Hongyu and Ni, Junfeng and Chen, Yixin and Li, Weishuo and Pei, Mingtao and Huang, Siyuan},
  booktitle=ICCV,
  year={2025}
}

@inproceedings{huang2024embodied,
  title={An Embodied Generalist Agent in 3D World},
  author={Huang, Jiangyong and Yong, Silong and Ma, Xiaojian and Linghu, Xiongkun and Li, Puhao and Wang, Yan and Li, Qing and Zhu, Song-Chun and Jia, Baoxiong and Huang, Siyuan},
  booktitle={Proceedings of the International Conference on Machine Learning (ICML)},
  year={2024}
}

@inproceedings{ni2024phyrecon,
  title={PhyRecon: Physically Plausible Neural Scene Reconstruction}, 
  author={Ni, Junfeng and Chen, Yixin and Jing, Bohan and Jiang, Nan and Wang, Bin and Dai, Bo and Li, Puhao and Zhu, Yixin and Zhu, Song-Chun and Huang, Siyuan},
  journal={Advances in Neural Information Processing Systems},
  year={2024}
}

@inproceedings{liu2025building,
  title={Building Interactable Replicas of Complex Articulated Objects via Gaussian Splatting},
  author={Liu, Yu and Jia, Baoxiong and Lu, Ruijie and Ni, Junfeng and Zhu, Song-Chun and Huang, Siyuan},
  booktitle={The Thirteenth International Conference on Learning Representations},
  year={2025},
}

@article{zhu2025mtu,
  title = {Move to Understand a 3D Scene: Bridging Visual Grounding and Exploration for Efficient and Versatile Embodied Navigation},
  author = {Zhu, Ziyu and Wang, Xilin and Li, Yixuan and Zhang, Zhuofan and Ma, Xiaojian and Chen, Yixin and Jia, Baoxiong and Liang, Wei and Yu, Qian and Deng, Zhidong and Huang, Siyuan and Li, Qing},
  journal = {International Conference on Computer Vision (ICCV)},
  year = {2025}  
}

@inproceedings{linghu2024msr3d,
  author    = {Linghu, Xiongkun and Huang, Jiangyong and Niu, Xuesong and Ma, Xiaojian and Jia, Baoxiong and Huang, Siyuan},
  booktitle = {Advances in Neural Information Processing Systems},
  editor    = {A. Globerson and L. Mackey and D. Belgrave and A. Fan and U. Paquet and J. Tomczak and C. Zhang},
  pages     = {140903--140936},
  publisher = {Curran Associates, Inc.},
  title     = {Multi-modal Situated Reasoning in 3D Scenes},
  url       = {https://proceedings.neurips.cc/paper_files/paper/2024/file/feaeec8ec2d3cb131fe18517ff14ec1f-Paper-Datasets_and_Benchmarks_Track.pdf},
  volume    = {37},
  year      = {2024}
}
}
\clearpage
% \maketitlesupplementary

% \input{X_suppl}

\end{document}